\documentclass[11pt]{article}

\usepackage{emnlp2021}

\usepackage{times}
\usepackage{latexsym}

\usepackage[T1]{fontenc}

\usepackage[utf8]{inputenc}

\usepackage{microtype}

\usepackage[utf8]{inputenc} 
\usepackage[T1]{fontenc}    
\usepackage{hyperref}       
\usepackage{url}            
\usepackage{booktabs}       
\usepackage{amsfonts}       
\usepackage{amsmath}
\usepackage{amssymb}
\usepackage{nicefrac}       
\usepackage{microtype}      
\usepackage{graphicx}
\usepackage{mathtools}
\usepackage{bbm}
\usepackage{amsthm}
\usepackage{tagging}
\usepackage{enumitem}

\newtheorem{prop}{Proposition}
\newtheorem{lemma}{Lemma}

\DeclarePairedDelimiter\abs{\lvert}{\rvert}%
\DeclarePairedDelimiter\norm{\lVert}{\rVert}%

\makeatletter
\let\oldabs\abs
\def\abs{\@ifstar{\oldabs}{\oldabs*}}
\let\oldnorm\norm
\def\norm{\@ifstar{\oldnorm}{\oldnorm*}}
\makeatother

\usepackage{xcolor}

\renewcommand{\comment}[1]{}

\newcommand{\exampletext}[1]{\emph{``#1''}}

\newcommand{\aspace}{\hspace{1em}}
\newcommand{\uci}{$^{\heartsuit}$}
\newcommand{\uw}{$^{\clubsuit}$}
\newcommand{\aiTwo}{$^{\spadesuit}$}

\title{Competency Problems:\\On Finding and Removing Artifacts in Language Data}

\author{Matt Gardner*\thanks{~~Equal contribution}\aiTwo \aspace
William Merrill*\aiTwo \aspace
Jesse Dodge\aiTwo \aspace\\
{\bf Matthew E. Peters}\aiTwo \aspace
{\bf Alexis Ross}\aiTwo \aspace
{\bf Sameer Singh}\uci \aspace
{\bf Noah A. Smith}\uw\aiTwo \aspace\\
\aiTwo Allen Institute for Artificial Intelligence\\
\uci University of California, Irvine\\
\uw University of Washington
} 

\begin{document}

\maketitle

\begin{abstract}

    Much recent work in NLP has documented dataset artifacts, bias, and spurious correlations between input features and output labels.  However, how to tell which features have ``spurious'' instead of legitimate correlations is typically left unspecified.  In this work we argue that for complex language understanding tasks, \emph{all} simple feature correlations are spurious, and we formalize this notion into a class of problems which we call \emph{competency problems}.  For example, the word ``amazing'' on its own should not give information about a sentiment label independent of the context in which it appears, which could include negation, metaphor, sarcasm, etc.  We theoretically analyze the difficulty of creating data for competency problems when human bias is taken into account, showing that realistic datasets will increasingly deviate from competency problems as dataset size increases.  This analysis gives us a simple statistical test for dataset artifacts, which we use to show more subtle biases than were described in prior work, including demonstrating that models are inappropriately affected by these less extreme biases.  Our theoretical treatment of this problem also allows us to analyze proposed solutions, such as making local edits to dataset instances, and to give recommendations for future data collection and model design efforts that target competency problems.

\end{abstract}

\section{Introduction}

Attempts by the natural language processing community to get machines to understand language or read text are often stymied in part by issues in our datasets~\citep{chen-etal-2016-thorough,sugawara-etal-2018-makes}.
Many recent papers have shown that popular datasets are prone to shortcuts, dataset artifacts, bias, and spurious correlations~\citep{jia-liang-2017-adversarial,rudinger-etal-2018-gender,ws-2019-gender}.  While these empirical demonstrations of deficiencies in the data are useful, they often leave unanswered fundamental questions of what exactly makes a correlation ``spurious'', instead of a feature that is legitimately predictive of some target label.

In this work we attempt to address this question theoretically.  We begin with the assumption that in a language understanding problem, no single feature on its own should contain information about the class label.  That is, \emph{all} simple correlations between input features and output labels are spurious: $p(y|x_i)$, for any feature $x_i$, should be uniform over the class label.  We call the class of problems that meet this assumption \emph{competency problems} (\S\ref{sec:competency}).\footnote{Our use of the term ``competency problems'' is inspired by, but not identical to, the term ``competence'' in linguistics.  We are referring to the notion that humans can understand essentially any well-formed utterance in their native language.}

\begin{figure}
    \centering
    \includegraphics[width=\columnwidth]{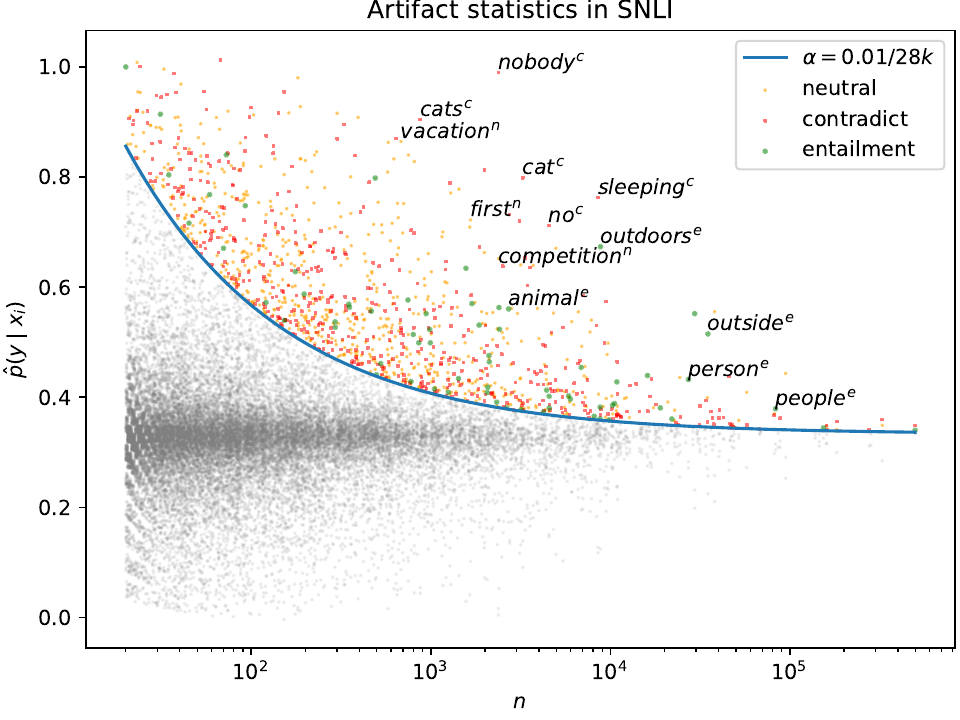}
\caption{A statistical test for deviation from a competency problem, where no individual feature (here words) should give information about the class label, plotting the number of occurrences of each word against the conditional probability of the label given the presence of the word.
The label associated with each point is marked by color and superscript.
All features above the blue line have detectable correlation with class labels, using a very conservative Bonferroni-corrected statistical test.}
    \label{fig:snli-artifacts}
\end{figure}

This assumption places a very strong restriction on the problems being studied, but we argue that it is a reasonable description of complex language understanding problems.  Consider, for example, the problem of sentiment analysis on movie reviews.  A single feature might be the presence of the word \exampletext{amazing}, which could be legitimately correlated with positive sentiment in some randomly-sampled collection of actual movie reviews.  However, that correlation tells us more about word frequency in movie reviews than it tells us about a machine's ability to understand the complexities of natural language.  A competent speaker of a natural language would know that \exampletext{amazing} can appear in many contexts that do not have positive sentiment and would not base their prediction on the presence of this feature alone.  That is, the information about the sentiment of a review, and indeed the \emph{meaning} of natural language, is contained in complex feature interactions, not in isolated features.  To evaluate a machine's understanding of language, we must remove all simple feature correlations that would allow the machine to predict the correct label without considering how those features interact.

Collecting data that accurately reflects the assumptions of a competency problem is very challenging, especially when humans are involved in creating it.  Humans suffer from many different kinds of bias and priming effects, which we collectively model in this work with rejection sampling during data collection.  We theoretically analyze data collection under this biased sampling process, showing that any amount of bias will result in increasing probability of statistically-significant spurious feature correlations as dataset size increases (\S\ref{sec:biased-sampling}).

This theoretical treatment of bias in data collection gives us a new, simple measure of data artifacts (\S\ref{sec:hypothesis}), which we use to explore artifacts in several existing datasets (\S\ref{sec:empirical}).  Figure~\ref{fig:snli-artifacts} revisits prior analyses on the SNLI dataset~\cite{bowman-etal-2015-large} with our statistical test.  An analysis based on pointwise mutual information~\citep[e.g.,][]{gururangan-etal-2018-annotation} would correspond to a horizontal line in that figure, missing many features that have less extreme but still significant correlations with class labels.  These less extreme correlations still lead models to overweight simple features.  The problem of bias in data collection is pervasive and not easily addressed with current learning techniques.

Our framework also allows us to examine the theoretical impact of proposed techniques to mitigate bias, including performing local edits after data collection (\S\ref{sec:local-edits}) and filtering collected data (\S\ref{sec:other-mitigations}).  We derive properties of any local edit procedure that must hold for the procedure to effectively remove data artifacts.   These proofs give dataset builders tools to monitor the data collection process to be sure that resultant datasets are as artifact-free as possible.  Our analysis of local edits additionally suggests a strong relationship to \emph{sensitivity} in boolean functions \citep{o2014analysis}, and we identify gaps in the theory of sensitivity that need to be filled to properly account for bias in sampled datasets.

We believe our theoretical analysis of these problems provides a good starting point for future analyses of methods to improve NLP data collection, as well as insights for inductive biases that could be introduced to better model competency problems.

\section{Competency Problems}
\label{sec:competency}

We define a \emph{competency problem} to be one where the marginal distribution over labels given any single feature is uniform.  For our analysis, we restrict ourselves to boolean functions: we assume an input vector $\mathbf{x}$ and an output value $y$, where $\mathbf{x} \in \{0, 1\}^d$ and $y \in \{0, 1\}$.\footnote{Boolean functions are quite general, and many machine learning problems can be framed this way. For NLP, consider that before the rise of embedding methods, language was often represented in machine learning models as bags of features in a very high-dimensional feature space, exactly as we are modeling the problem here. The first (embedding) layer of a modern transformer is still very similar to this, with the addition of a position encoding. The choice of what counts as a ``simple feature'' is admittedly somewhat arbitrary; we believe that considering word types as simple features, as we do in most of our analysis, is uncontroversial, but there are other more complex features which one still might want to control for in competency problems.} In this setting, competency means $p(y|x_i) = 0.5$  for all $i$.  In other words, the information mapping $\mathbf{x}$ to $y$ is found in complex feature interactions, not in individual features.

Our core claim is that language understanding requires composing together many pieces of meaning, each of which on its own is largely uninformative about the meaning of the whole.  We do not believe this claim is controversial or new, but its implications for posing language understanding as a machine learning problem are underappreciated and somewhat counterintuitive.  If a model picks up on individual feature correlations in a dataset, it has learned something \emph{extra-linguistic}, such as information about human biases, not about how words come together to form meaning, which is the heart of natural language understanding.  To push machines towards linguistic competence, we must control for all sources of extra-linguistic information, ensuring that no simple features contain information about class labels.

For some language understanding problems, such as natural language inference, this intuition is already widely held.  We find it surprising and problematic when the presence of the word \exampletext{cat}, \exampletext{sleeping} or even \exampletext{not} in either the premise or the hypothesis gives a strong signal about an entailment decision~\citep{gururangan-etal-2018-annotation,poliak-etal-2018-hypothesis}.  Competency problems are broader than this, however.  Consider the case of sentiment analysis.  It is true that a movie review containing the word \exampletext{amazing} is more likely than not to express positive sentiment about the movie.  This is because of distributional effects in how humans choose to use phrases in movie reviews.  These distributional effects cause the \emph{lexical semantics} of \exampletext{amazing} to carry over into the whole context, essentially conflating lexical and contextual cues.  If our goal is to build a system that can accurately classify the sentiment of movie reviews, exploiting this conflation is useful.  But if our goal is instead to build a machine that understands how sentiment is expressed in language, this feature is a red herring that must be controlled for to truly test linguistic competence.

\section{Biased Sampling}
\label{sec:biased-sampling}

To get machines to perform well on competency problems, we need data that accurately reflects the competency assumption, both to evaluate systems and (presumably) to train them.  However, humans suffer from blind spots, social bias, priming, and other psychological effects that make collecting data for competency problems challenging.  Examples of these effects include instructions in a crowdsourcing task that prime workers to use particular language,\footnote{This is ubiquitous in crowdsourcing; see, e.g., common patterns in DROP~\citep{dua-etal-2019-drop} or ROPES~\citep{lin-etal-2019-reasoning} that ultimately derive from annotator instructions.} or distributional effects in source material, such as the \exampletext{amazing} examples above, or racial bias in face recognition~\citep{Buolamwini2018GenderSI} and abusive language detection datasets~\citep{davidson-etal-2019-racial, sap2019risk}.

In order to formally analyze the impact of human bias on collecting data for competency problems, we need a plausible model of this bias.  We represent bias as rejection sampling from the target competency distribution based on single feature values.  Specifically, we assume the following dataset collection procedure.  First, a person samples an instance from an unbiased distribution $p_u(\mathbf{x}, y)$ where the competency assumption holds.  The person examines this instance, and if feature $x_i = 1$ appears with label $y = 0$, the person rejects the instance and samples a new one, with probability $r_i$.  If $y = 0$ corresponds to negative sentiment and $x_i$ indicates the presence of the word \exampletext{amazing}, a high value for $r_i$ would lead to \exampletext{amazing} appearing more often with positive sentiment, as is observed in typical sentiment analysis datasets.

We do not that claim rejection sampling is a plausible psychological model of dataset construction.  However, we \emph{do} think it is a reasonable first-order approximation of the \emph{outcome} of human bias on data creation, for a broad class of biases that have empirically been found in existing datasets, and it is relatively easy to analyze.

\subsection{Emergence of Artifacts Under Rejection Sampling}
\label{sec:rejection-sampling}

Let $p_u(y|x_i)$ be the conditional probability of $y=1$ given $x_i=1$ under the unbiased distribution, $p_b(y|x_i)$ be the same probability under the biased distribution, and $\hat{p}(y|x_i)$ denote the empirical probability within a biased dataset of $n$ samples. Additionally, let $f_i$ be the marginal probability $p_u(x_i)$.  Recall that $p_u(y|x_i)$ is $0.5$ by assumption.

We will say that dimension $i$ has an \emph{artifact} if the empirical probability $\hat{p}(y|x_i)$ statistically differs from $0.5$. In this section, we will show that an artifact emerges if there is a bias at dimension $i$ in the sampling procedure, which is inevitable for some features in practice. We will formalize this bias in terms of a rejection sampling probability $r_i$.

For a single sample $\mathbf{x}, y$, we first derive the joint and marginal probabilities $p_b(y, x_i)$ and $p_b(x_i)$, from which we can obtain $p_b(y|x_i)$.  These formulas use a recurrence relation obtained from the rejection sampling procedure.
\begin{align*}
    p_b(y, x_i) &= \frac{1}{2} f_i + \frac{1}{2} f_i r_i p_b(y, x_i) \\
    \therefore p_b(y, x_i) &= \frac{f_i}{2 - f_i r_i} \\
\end{align*}
\begin{align*}
    p_b(x_i) &= \frac{1}{2} f_i + \frac{1}{2} f_i (1 - r_i) + \frac{1}{2} f_i r_i p_b(x_i) \\
    \therefore p_b(x_i) &= \frac{2 f_i - f_i r_i}{2 - f_i r_i} \\
    \therefore p_b(y \mid x_i) &= \frac{p_b(y, x_i)}{p_b(x_i)} = \frac{1}{2 - r_i}
\end{align*}
With no bias ($r_i = 0$), this probability is $0.5$, as expected, and it rises to $1$ as $r_i$ increases to $1$.


We define $\hat{p}(y|x_i)$ as the empirical expectation of $p_b(y|x_i)$ over $n$ samples containing $x_i$, with different samples indexed by superscript $j$.
$\hat{p}(y|x_i) = \frac{1}{n} \sum_{j=1}^n y^j$.
Note that $\hat{p}$ is a conditional binomial random variable. By the central limit theorem, $\hat{p}$ is approximately $\sim \mathcal N(\mu_{\hat{p}}, \sigma_{\hat{p}}^2)$ for large $n$, where
\begin{align*}
    \mu_{\hat{p}} &= p_b(y \mid x_i) = \frac{1}{2 - r_i} \\
    \sigma_{\hat{p}}^2 &= \left( \frac{1 - r_i}{(2 - r_i)^2} \right)^2 \cdot \frac{1}{n} .
\end{align*}

This variance is inversely proportional to the number of samples $n$. Thus, $\hat{p}(y|x_i)$ can be well approximated by its expected value for a large number of samples.
As the rejection probability $r_i$ increases, the center of this distribution tends from $0.5$ to $1$. This formalizes the idea that bias in the sampling procedure will cause the empirical probability $\hat{p}( y|x_i )$ to deviate from $0.5$, even if the ``true'' probability is $0.5$ by assumption. Increasing the sample size $n$ concentrates the distribution inversely proportional to $\sqrt{n}$, but the expected value is unchanged.
Thus, artifacts created by rejection sampling will not be combated by simply sampling more data from the same biased procedure---the empirical probability will still be biased by $r_i$ even if $n$ increases arbitrarily.
These persistent artifacts can be exploited at \emph{i.i.d.} test time to achieve high performance, but will necessarily fail if the learner is evaluated under the competency setting.

\subsection{Hypothesis Test}
\label{sec:hypothesis}

Here we set up a hypothesis test to evaluate if there is enough evidence to reject the hypothesis that $r_i$ is 0, i.e., that the data is unbiased. In this case, we can use a one-sided binomial proportion hypothesis test, as our rejection sampling can only lead to binomial proportions for $p_b(y\mid x_i)$ that are greater than $\frac{1}{2}$. Our null hypothesis is that the binomial proportion $p_b(y\mid x_i)=0.5=p_0$, or equivalently, that $r_i=0$. Our alternative hypothesis is that $p_b(y\mid x_i)\geq 0.5$. Let $\hat{p}$ be the observed probability. We can compute a $z$-statistic\iftagged{shortened}{}{\footnote{The use of a $z$-statistic depends on the normal approximation to a binomial distribution, which holds for large $n$.}} using the standard formula:
\begin{align}
    \label{eq:z-score}
    z^* = \frac{\hat{p}-p_0}{\sqrt{p_0(1-p_0)/n}}
\end{align}
Thus, if our observed proportion $\hat{p}$ is far from $p_0=0.5$, we will have enough evidence to reject the null hypothesis that $r_i=0$. This depends on $n$ as well, and to explore this interaction, we solve for $\hat{p}$ for a given $n$ and confidence level $z^*$: $\hat{p} = \frac{z^*}{2\sqrt{n}}+\frac{1}{2}$.

\section{Empirical Analysis}
\label{sec:empirical}

With a hypothesis test in hand, we can examine existing datasets for evidence of statistically-significant feature bias, and then explore the extent to which this bias impacts models supervised with this data.  Prior work has used pointwise mutual information (PMI) to find features that have high correlation with labels~\citep[e.g.,][]{gururangan-etal-2018-annotation}.  This measure is useful for understanding why certain features might get used as deterministic decision rules by models~\citep{Ribeiro2018AnchorsHM,wallace-etal-2019-universal}.  However, studies involving PMI have also intuitively understood that PMI by itself does not tell the whole story, as a strict ranking by PMI would return features that only appear once in the dataset.  To account for this problem, they used arbitrary cutoffs and included information about feature occurrence in addition to their PMI ranking.  A benefit of our approach to defining and detecting artifacts is that we have a single statistical test that takes into account both the number of times a feature appears and how correlated it is with a single label.  We use this test to find features with the strongest statistical evidence for artifacts (\S\ref{sec:empirical-data}) and then show empirically that models use these features inappropriately when making predictions (\S\ref{sec:empirical-model}).  This analysis goes beyond deterministic prediction rules, showing that the impact of sampling bias on model behavior is subtle and pervasive.

\subsection{Data Analysis}
\label{sec:empirical-data}
We analyze two datasets with the hypothesis test from \S\ref{sec:hypothesis}: SNLI~\cite{bowman-etal-2015-large} and the Universal Dependencies English Web Treebank~\cite{silveira14gold}.

\paragraph{SNLI} 
Each feature $x_i$ represents the presence of a word in a given example, counting each appearance in an instance as a separate occurrence\footnote{We remove punctuation and tokenize on whitespace only.} for the purposes of computing $n$ and $\hat{p}$ in Equation~\ref{eq:z-score}. We compute a $z$-statistic for every token that appears in the SNLI data, where $p_0 = \frac{1}{3}$, as SNLI has three labels.  
We then plot the $z$-statistic for each token against the number of times the token appears in the data.  We also plot a curve for the value of the $z$-statistic at which the null hypothesis (that $r_i = 0$) should be rejected, using a significance level of $\alpha = 0.01$ and a conservative Bonferroni correction~\cite{bonferroni-correction} for all 28,000 vocabulary items.  This analysis is shown in Figure~\ref{fig:snli-artifacts}.  We label in Figure~\ref{fig:snli-artifacts} several words that were also found to be artifacts by \citet{gururangan-etal-2018-annotation} and \citet{wallace-etal-2019-universal}, among others.

We find a very large number of deviations from the competency assumption, many more than would be suggested by a PMI-based analysis.  PMI equals $\log\frac{\hat{p}(y|x_i)}{\hat{p}(y)}$; because $\hat{p}(y)$ does not vary across features, and the data is balanced over labels, a PMI analysis ranks features by $\hat{p}(y|x_i)$, looking only at the $y$-axis in Figure~\ref{fig:snli-artifacts}.\iftagged{shortened}{}{\footnote{In practice an arbitrary threshold is chosen on the $x$-axis to avoid rare features.  Again here a statistical test is a more principled way to account for rare features.}}  But the threshold for which a deviation in $\hat{p}(y|x_i)$ becomes a statistical artifact depends
on the number of times the feature is seen, so our statistical test gives a simpler and more complete picture of data artifacts.  Strong statistical deviations with less extreme PMI values still impact model behavior (\S\ref{sec:empirical-model} and \S\ref{sec:other-mitigations}).

\paragraph{UD English Web Treebank} Next we turn to dependency parsing. In particular, we focus on the classic problem of prepositional phrase (PP) attachment \cite{collins-brooks-1995-prepositional}, which involves determining whether a PP attaches to a verb (e.g., \textit{We ate spaghetti with forks}) or a noun (e.g., \textit{We ate spaghetti with meatballs}). We heuristically extract (verb, noun, prepositional phrase) constructions with ambiguous attachment from the UD English Web Treebank (EWT) training data.\footnote{See Appendix \ref{sec:tuple-extraction} for how we extract these constructions.} We treat (verb, preposition) tuples as features and attachment types (\textit{noun} or \textit{verb}) as labels, and we compute a $z$-statistic for each tuple.

Figure \ref{fig:pp-attach} shows the $z$-statistic for each tuple that appears 10 or more times in the data. We labeled tuples that also appear in the locally edited samples from the UD English contrast set created by \citet{gardner-etal-2020-evaluating}. Many of these tuples fall either above or close to the significance curve, suggesting that the low contrast consistency reported by \citet{gardner-etal-2020-evaluating} could potentially be explained by models' reliance on these artifacts.\footnote{Some tuples in the plot with high $p(y|x_i)$ are not artifacts, as the attachment decision is nearly deterministic. For instance, the top right blue dot corresponds to (\textit{have, of}); \textit{of} can only attach to \textit{have} in archaic or idiosyncratic constructions.} 

\begin{figure}
    \centering
    \includegraphics[width=.8\columnwidth]{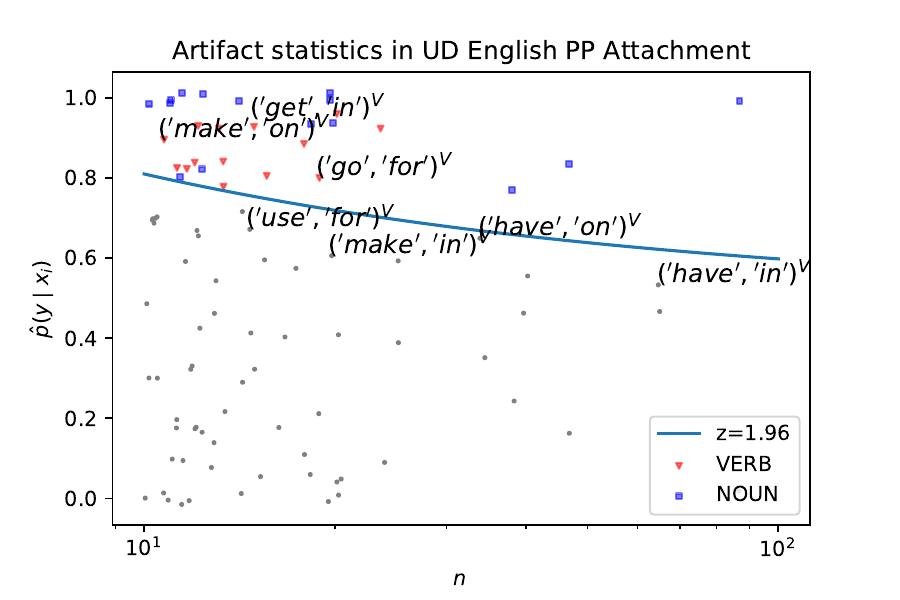}
    \caption{
    Artifact statistics of (verb, preposition) tuples in PP attachment in the UD English Web Treebank (EWT). Plotted are tuples that appear in the original EWT training data, and labeled are tuples that also appear in the UD English locally edited contrast set.}
    \label{fig:pp-attach}
\end{figure}

\subsection{Model Analysis}
\label{sec:empirical-model}

The previous section reveals a large number of individual word artifacts in the SNLI dataset.  Here, we ask whether typical NLP models learn to bias their predictions based on these artifacts for both the SNLI and RTE \cite{Dagan2005ThePR}\footnote{We use the RTE data from SuperGlue \cite{Wang2019SuperGLUEAS}.} datasets.  That is, we will show that these single words noticeably influence a model's confidence in particular predictions, even when the PMI value is not extreme enough to create a universal trigger~\citep{wallace-etal-2019-universal}.  Importantly, this analysis focuses on words with high $z$-statistics, which are often words that show up very frequently with slight deviations from $p_u(y|x_i)$.  This includes words such as \exampletext{for} and \exampletext{to} (the two words with highest $z$-statistic for the neutral class), and \exampletext{there} and \exampletext{near} (the highest and fifth-highest $z$-statistic for the entailment class). 

To measure the model bias learned from these words, we employ RoBERTa-base \cite{Liu2019RoBERTaAR} fine-tuned on RTE, and ALBERT-base \cite{Lan2020ALBERTAL} fine-tuned on SNLI.\footnote{Both models are from \citet{morris2020textattack}, and are implemented in the Transformers library \cite{wolf-etal-2020-transformers}.}  Given a single type such as $x_i=$ \exampletext{nobody} and a target class such as $y=$ ``contradiction'', we estimate the model $\tilde{p}(y|x_i)$ as follows.  We first create two synthetic input examples, one with the premise containing only the single token with an empty hypothesis, and one with an empty premise and hypothesis containing the single token.  As each input contains only a single token without additional context, this tests whether the model will bias its output based on the token.  We run a forward pass with each input and average the target class probabilities as an estimate of $\tilde{p}(y|x_i)$.  All of the words in each dataset appearing at least 20 times 
are partitioned among the classes based on their largest class conditional $z^*$, and for each class we form two cohorts of 50 words each with the highest and lowest $z^*$.
Let $\mathcal{X}_{y,z^*_<}$ denote the set of $x_i$ with the lowest $z^*$ for class $y$ and similarly let $\mathcal{X}_{y,z^*_>}$ denote the set with the largest $z^*$.
Finally, we compute the average $\Delta \tilde{p}_y = \sum_{x_i \in \mathcal{X}_{y,z^*_>}} \tilde{p}(y|x_i) - \sum_{x_i \in \mathcal{X}_{y,z^*_<}} \tilde{p}(y|x_i)$.

The results are shown in Table \ref{table:model_bias}.  As can be seen, these models exhibit non-trivial bias based on the single token inputs, with $\Delta \tilde{p}_y$ exceeding 10\% for some classes.  The bias is much more extreme for SNLI versus RTE, likely due to the fact that RTE has two orders of magnitude less data than SNLI.

A caveat about this experiment is in order: due to the fact that automatically replacing high $z^*$ words with low $z^*$ words will likely make most inputs nonsensical, we chose to use very unnatural single-word inputs to the model instead.  We believe this is a reasonable estimate of the model's marginal prior on these tokens, measured in a way that introduces the fewest possible confounding variables into the experiment, but it's possible that it does not completely reflect how a model treats these tokens in context.  Section \ref{sec:other-mitigations} discusses some additional empirical evidence for models' reliance on these artifacts.

\begin{table}
\centering
\small
\begin{tabular}{llr}
\toprule
Dataset & Class & $\Delta \hat{p}_y$ \\
\midrule
RTE & entailment & +2.2 \% \\
SNLI & entailment & +14.7 \% \\
SNLI & neutral & +7.9 \% \\
SNLI & contradiction & +12.5 \%


\end{tabular}
\caption{Comparison of the average model predicted class probability for a single token across two cohorts of large $z^*$ tokens and low $z^*$ tokens using RoBERTa (RTE) and ALBERT (SNLI).}
\label{table:model_bias}
\end{table}

\section{Mitigating Artifacts with Local Edits}
\label{sec:local-edits}

Many works have tried to remove data artifacts by making minimal changes to existing data~\cite[\emph{inter alia}]{shekhar2017foil,sennrich2016grammatical,zhao-etal-2018-gender}.  In this section we show that this kind of data augmentation can be effective with an appropriately sensitive edit model, where \emph{sensitivity} refers to how often a change to inputs results in the label changing.  However, because humans are involved in making these changes, achieving appropriate sensitivity is challenging, and bias in this process can lead to the introduction of new artifacts.  This suggests that care must be taken when performing edit-based data augmentation, as large edited training datasets are not likely to be artifact-free~\citep[cf.][]{tafjord-etal-2019-quartz,huang-etal-2020-counterfactually}.

Imagine a new dataset $\mathcal{D}_e$ consisting of samples $\mathbf{x}', y'$ generated by making local edits according to the following repeated procedure:
\begin{enumerate}[nosep]
    \item Randomly sample an instance $\mathbf{x}$ from a dataset $\mathcal{D}_b$ of $n$ instances created under $p_b$. 
    \item Make some changes to $\mathbf{x}$ to arrive at $\mathbf{x'}$.
    \item Manually label $y'$ and add $\langle \mathbf{x'}, y'\rangle$ to $\mathcal{D}_e$.
\end{enumerate}
We examine the expected probability $p_e(y'|x'_i)$ under this edit process.  
Informally, this probability should depend on how often a change to $\mathbf{x}$ affects $y$. Formalizing this, we define the \emph{edit sensitivity} $s_i$ to be the probability that $y$ changes during editing given the occurrence of a particular feature in the edited data, i.e.,
\begin{equation*}
    s_i = p_b(y' = \neg y \mid x_i') .
\end{equation*}
The other quantity of interest for an edit model is $e_i$, the probability that dimension $i$ gets flipped when going from $\mathbf{x}$ to $\mathbf{x'}$.
In order to make theoretical progress, we also need to make strong independence assumptions on $x_i$, $e_i$ and $s_i$; we will examine these assumptions momentarily.  We first show that under these assumptions, $s_i$ and $e_i$ control whether samples generated by local editing debias first-order artifacts.

\begin{prop}[Proof in \S\ref{sec:prop1-proof}] \label{thm:edits}
\textit{Assume $x_i$, $x_j$, $e_i$, $e_j$, $s_i$, and $s_j$ are independent for all $i$, $j$.
Then $p_e(y' \mid x'_i) = \frac{1}{2}$ if and only if $r_i = 0$ or $s_i = \frac{1}{2}$ or $e_i = 1$.}
\end{prop}
This proposition shows that there are three ways to achieve unbiased data from a local edit procedure that edits dimensions independently: (1) start with unbiased data, (2) always flip every feature, and (3) flip the label half the time for each feature.

The first of these conditions is not under the control of the edit procedure, and if we start with unbiased data there is no need for debiasing.

The second condition is roughly analogous to the approach taken by most prior work that performs local edits: they aim to always change whichever features led to a prediction, with $s_i = 1$ and $e_i = 1$.  However, $e_i$ just indicates whether a feature was \emph{flipped}, which means that to achieve $e_i = 1$, every non-zero feature must be added to every instance where it is missing, which is somewhat nonsensical for language data, and thus this solution isn't practical.

This leaves the third condition as the only practical solution for local edit procedures to have a hope of debiasing datasets.  That is, these procedures should aim to flip the label on average half of the time, for each feature that is changed.  

We emphasize here that the assumptions we made in deriving this result are very strong.  If these assumptions are violated, it is easy to construct an adversarial edit procedure that will introduce bias into an unbiased dataset (the $r_i = 0$ solution).  Similarly, if $e_i$ and $s_i$ are correlated, one can construct cases that break the $s_i = \frac{1}{2}$ solution as well.

Furthermore, these independence assumptions are not realistically achievable for any human-produced local edits on language data.  We thus take the guarantees we derived with some skepticism, and view this result more as guidelines for how to set up and monitor local edit procedures: aim to flip the label roughly half the time, in a way that is uncorrelated with which features are getting edited, and monitor the resulting data for edit sensitivity and artifact statistics.  In the next section we show how to use the theoretical lens we have developed to analyze local edits that have been made in prior work.

\subsection{Local Edits in Practice}

We empirically investigate the effectiveness of local edits for reducing single feature artifacts using locally edited samples generated from two datasets: (1) the Boolean Questions dataset \citep[BoolQ;][]{clark2019boolq}, which consists of pairs of paragraphs ($p$) and questions ($q$), where each $q$ has a binary answer that can be found by reasoning over $p$; and (2) IMDb~\citep{maas-etal-2011-learning}, a sentiment classification dataset in the domain of movie reviews. We define each feature $x_i$ as the occurrence of a particular word within $q$ for BoolQ, and within the text of the review for IMDb. \citet{gardner-etal-2020-evaluating} generated additional data for BoolQ and IMDb by making local edits to the question or review text and recording the updated binary label.

\autoref{fig:boolq} visualizes the effect of these changes on single-feature artifacts by comparing the artifact statistics for the original texts to the statistics for the edited texts generated by \citet{gardner-etal-2020-evaluating}. For BoolQ, many tokens in the original data exhibit artifacts in the positive ($> 0.5$) direction, while, within the edited data, almost all tokens fall within the confidence region.  In contrast, there is no apparent distributional difference between artifact statistics for the original vs.~edited texts on IMDb.
We find that for BoolQ, the per-token edit sensitivity distribution has a median of $\tilde{s} = 0.429$ (mean $0.348$, std $0.274$), which, by \autoref{thm:edits}, explains why most of the $\hat p(y|x_i)$ values for the edited samples are not significantly different from $0.5$. For IMDb, $\tilde{s} = 1.00$ (mean $1$, std $0$).
This case study illustrates the importance of leveraging our theory to engineer better edit models.

\begin{figure}
    \centering
    \includegraphics[width=.8\columnwidth]{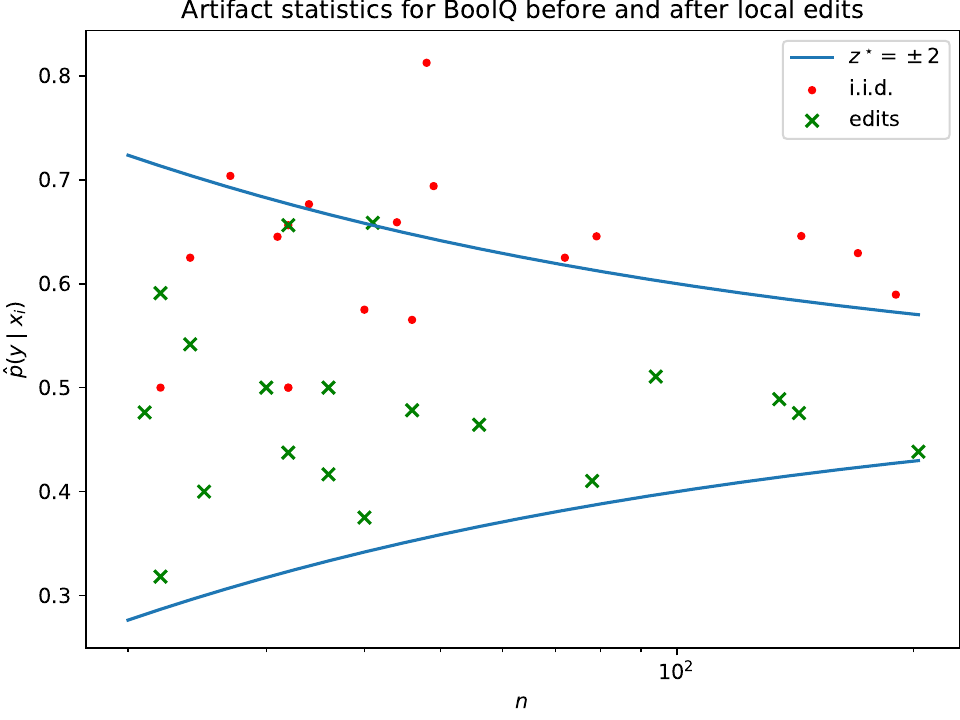}
    \includegraphics[width=.8\columnwidth]{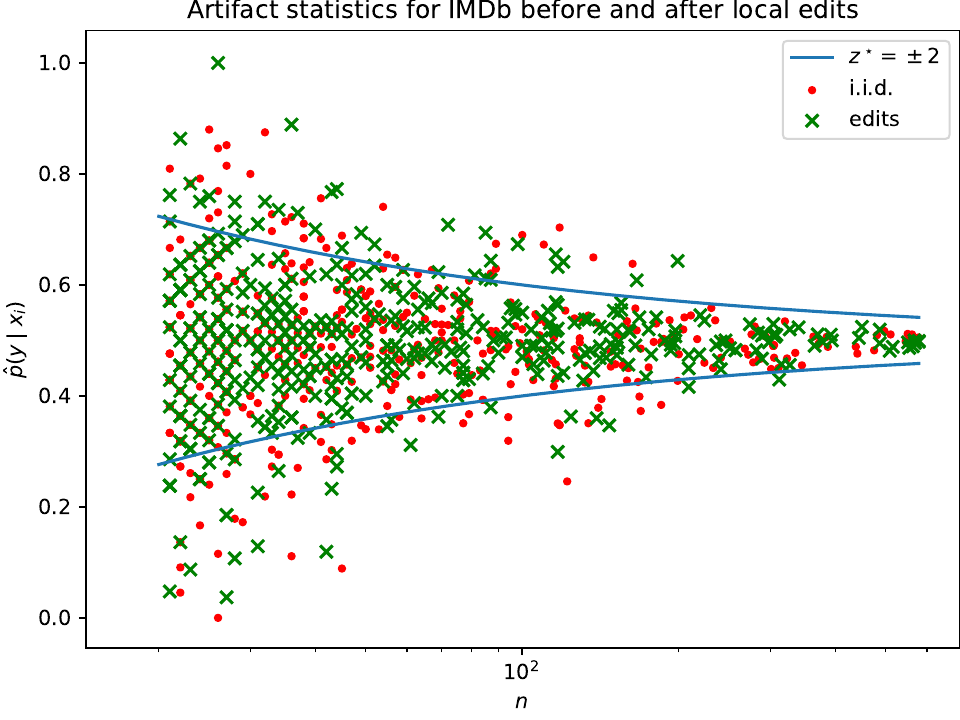}
    \caption{The artifact statistics of the original BoolQ (above) and IMDb (below) samples are plotted in red, compared to the artifact statistics over the edited instances, plotted in green.}
    \label{fig:boolq}
\end{figure}

\subsection{Local Edits and Boolean Sensitivity}

In the above discussion we used the term \emph{sensitivity} in an informal way to describe the probability that a local edit changes the label.  This term also has a related formal definition in the study of boolean functions, where it is an implicit complexity measure~\citep{wegener1987complexity}.  Sensitivity in this sense has been shown to correlate with generalization in neural networks~\citep{franco2001measure}, and has been extended for use with practical NLP datasets~\citep{hahn2021}.  In this section we discuss the intersection of our theory with sensitivity analysis, highlighting limitations in sensitivity analysis for sampled datasets that could be addressed in future work.

For a boolean vector $x$, let $x^i$ be the \emph{Hamming neighbor} of $x$ at dimension $i$: i.e., the vector where $x_i$ has been flipped and all other bits remain the same. Consider $f : \{0, 1\}^d \to \{0, 1\}$. The \emph{sensitivity set} $S(f, x)$ is the set of Hamming neighbors of $x$ with different labels:
\begin{equation*}
    S(f, x) = \left \{ i \in [0,\dots,d] \mid f(x) \neq f(x^i) \right \} .
\end{equation*}
The \emph{local sensitivity} $s(f, x)$ is the size of this set: $s(f, x) = \abs{S(f, x)}$. Finally, the \emph{global sensitivity} is defined as $s(f) = \max_x s(f, x)$.

\paragraph{Importance of sensitivity}
In our case, the effect of local editing on a dataset can be understood in terms of sensitivity. Imagine a boolean function $f : \{0, 1\}^d \to \{0, 1\}$ from which we draw $n$ samples $\langle \mathbf{x}, y\rangle$. If these samples are drawn uniformly over $\{0, 1\}^d$, then the probability of observing any Hamming neighbors goes to $0$ rapidly with $d$.\iftagged{shortened}{}{\footnote{This is essentially the curse of dimensionality.}}
Thus, it is possible to pick a low sensitivity function
that can perfectly fit the data. In this sense, the true sensitivity of $f$ is likely \emph{underspecified} by the dataset.

Imagine we give this data to a learner with inductive bias resembling some variant of Occam's razor. If the learner's notion of complexity is correlated with sensitivity (which many complexity measures are), then the learner will favor low sensitivity decision boundaries.  Thus, the fact that sensitivity is underspecified in the training data is a problem if the gold-standard function has high sensitivity, as the inductive bias of the learning algorithm may favor low-sensitivity alternatives.

Contrast this with a dataset where some local neighborhoods in the input space have been filled in with local edits. The set of observed neighbors around a point $x$ provide a lower bound on $s(f, x)$, which is a lower bound on $s(f)$. In this sense, $s(f)$ is no longer underspecified by the dataset. 

In this discussion we have used \emph{underspecified} in an informal way; there is no precise measure of the sensitivity of a sampled dataset (as opposed to a fully-specified function), particularly when generalizing from finite boolean functions to natural language inputs.  Attempts to generalize sensitivity to natural language have done so by leveraging large language models to generate neighbors from which sensitivity can be estimated~\citep{hahn2021}.  Resampling data in this way can give reasonable estimates of the sensitivity of the underlying task, but it is fundamentally incompatible with measuring dataset artifacts of the kind we discuss in this paper, as the generative model can fill in parts of the data distribution that are missing due to sampling bias, giving a higher estimate of sensitivity than is warranted by the sampled dataset.

\iftagged{shortened}{
\section{Data Filtering}
\label{sec:other-mitigations}
}{
\section{Other Mitigation Techniques}
\label{sec:other-mitigations}
In this section we briefly discuss the implications of our theoretical analysis for other artifact mitigation techniques that have been proposed in the literature.  Our analysis in this section is not rigorous and is meant only to give high-level intuition or potential starting points for future work.

\paragraph{More annotators}
One suggested mitigation technique for dataset artifacts is to increase the number of annotators~\cite{geva-etal-2019-modeling}.  Especially when people generate the text that is used in a dataset, there can be substantial person-specific correlations between features and labels.  Having more annotators washes out those correlations in aggregate, making the data less biased overall.

We briefly analyze this procedure using our rejection sampling framework.  For simplicity, we have so far only considered a single possible rejection probability, where an instance is rejected with probability $r_i$ if $x_i = 1$ and $y = 0$.  If we introduce additional rejection probabilities for the other three possible combinations of values for $x_i$ and $y$, there will be the possibility that some rejections balance out other rejections.  We can model multiple annotators by splitting a dataset into $k$ different slices that have their own bias vectors $\mathbf{r}$.  If the $\mathbf{r}$ vectors are uncorrelated, it seems likely that as $k$ increases, the probability that $\hat{p}(y|x_i)$ deviates from $p_u(y|x_i)$ tends towards zero.  Even in our simplistic model, if we assume a sparse $\mathbf{r}$, averaging more and more of them will make the deviation tend toward zero, if the non-zero dimensions are uncorrelated.

However, if the $\mathbf{r}$ vectors are correlated, increasing the number of annotators will not produce data reflecting the competency assumption.  When might the $\mathbf{r}$ vectors be correlated?  This could happen due to societal biases, word usage frequencies, or priming effects from data collection instructions given to all annotators.  Surely across any pool of annotators there will be some dimensions along which $\mathbf{r}$ values are correlated, and other dimensions along with they are not.  Increasing the number of annotators thus helps mitigate the problem, but does not solve it completely.

\paragraph{Data filtering}
}
A recent trend is to remove data from a training set that is biased in some way in order to get a model that generalizes better~\cite{Bras2020AdversarialFO,swayamdipta-etal-2020-dataset,oren-etal-2020-improving}.  While this method can be effective for very biased datasets, it is somewhat unsatisfying to remove entire instances because of bias in a single feature.  In the extreme case where $r_i \approx 1$, such as with \exampletext{nobody} in SNLI (Fig.~\ref{fig:snli-artifacts}), this process could effectively remove $x_i$ from the observed feature space.

To understand the effect of these automated methods on dataset artifacts, we repeat the analysis from \S\ref{sec:empirical-data} on data that was classified as ``ambiguous'' according to Dataset Cartography~\citep{swayamdipta-etal-2020-dataset}.  This data was shown to provide better generalization when used as training data compared to the original training set.  The ambiguous instances did not have a balanced label distribution, so we downsampled the data to balance it, then downsampled the whole training data to get the same number of instances as the balanced ambiguous set.

The resulting artifact plots are shown in Figure~\ref{fig:cartography-artifacts}.  As can be seen, the ``ambiguous'' instances have many fewer deviations from the competency assumption, across the entire range of our hypothesis test.  It is not just high PMI values that are getting corrected by finding ambiguous instances; all statistical deviations are impacted.  This effect is striking, and it further corroborates our arguments about the importance of the competency assumption.\footnote{Comparing the lower part of Figure~\ref{fig:cartography-artifacts} to Figure~\ref{fig:snli-artifacts} also corroborates our derived result (\S\ref{sec:rejection-sampling}) that larger datasets are more likely to have artifacts.  With 24\% of the data there are many fewer artifacts.}

\begin{figure}
    \centering
    \includegraphics[width=.8\columnwidth]{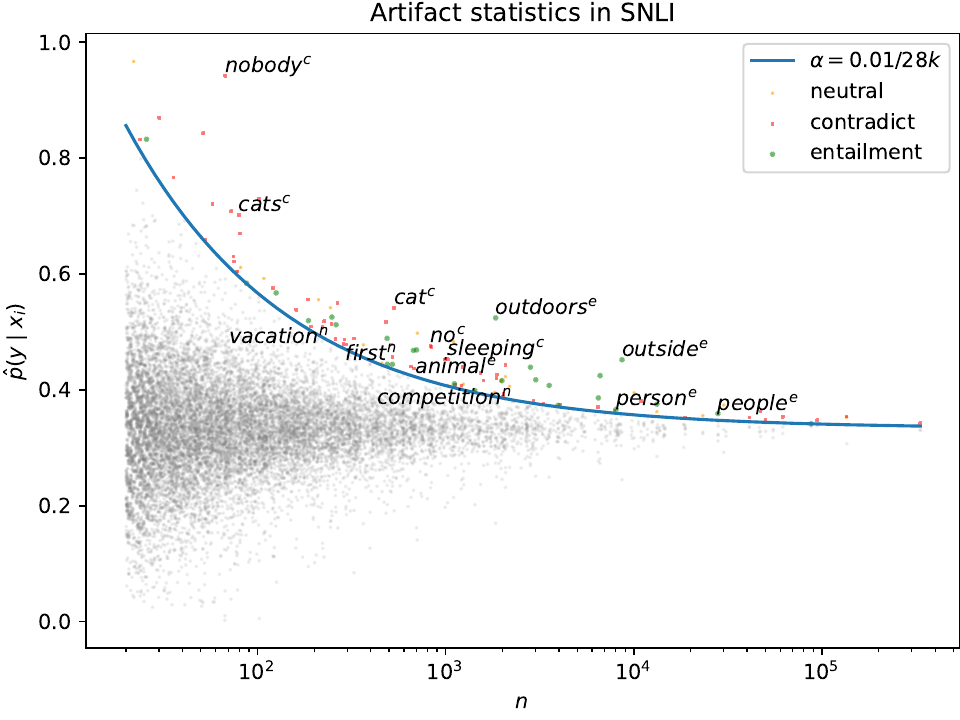}
    \includegraphics[width=.8\columnwidth]{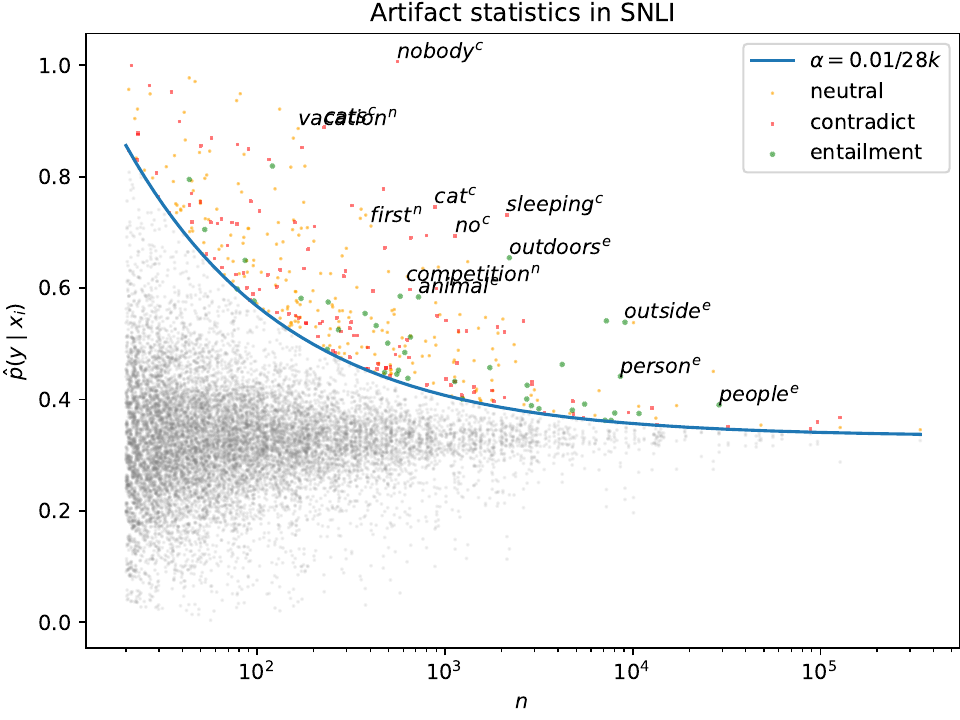}
    \caption{Statistical artifacts in ambiguous instances~(\citealp{swayamdipta-etal-2020-dataset}; above) versus a random (same-size) sample from the SNLI training set (below).  The filtering done by ambiguous instance detection targets statistical artifacts across the whole range of the statistical test, not just high PMI values.}
    \label{fig:cartography-artifacts}
\end{figure}

\section{Other Related Work}

\paragraph{Theoretical analysis of bias}
Several recent works explore sources and theoretical treatments of bias or spurious correlations in NLP~\cite{shah-etal-2020-predictive,Kaushik2020ExplainingTE} or ML more broadly~\cite{shah-2020-simplicity-bias}.  Our work differs by introducing a competency assumption and exploring its implications.  The difference between our biased and unbiased distributions is an instance of covariate shift~\cite{QuioneroCandela2009DatasetSI}.

\paragraph{Competent models}
An interesting question is whether we can inject a ``competency inductive bias'' into models, i.e., discourage relying on individual features.
The closest works we are aware of are methods that ensemble weak models together with strong models during training~\citep{clark-etal-2020-learning,dagaev2021toogoodtobetrue}, or ensembles of models with unaligned gradients~\citep{teney2021evading}.
\iftagged{shortened}{}{Other works use ensembles with models targeted at known sources of data artifacts, but these are less close to a competency assumption~\cite{clark-etal-2019-dont,karimi-mahabadi-etal-2020-end}.}

\section{Conclusion}
The more NLP models advance, the better they are at learning statistical patterns in datasets.  This is problematic for language understanding research if some statistical patterns allow a model to bypass linguistic competence.  We have formalized this intuition with a class of problems called \emph{competency problems}, arguing that, for any language understanding task, \emph{all} correlations between simple features and labels are spurious.  Collecting data meeting this assumption is challenging, but we have provided theoretical analysis that can inform future data collection efforts for such tasks.

We conclude with some final thoughts on general best practices for data collection, informed by the analysis in this paper.  If annotators are generating text for some data collection task, find ways to decrease priming effects.  This could involve using images as prompts instead of text~\cite{novikova-etal-2017-e2e,weller-etal-2020-learning}, or randomly sampling words to include in the generated text.  If existing text is being collected and annotated, make local edits to the text while monitoring the sensitivity of those edits according to the guidelines in \S\ref{sec:local-edits}, perhaps using different processes between train and test, to minimize correlations between train features and test labels.

\section*{Acknowledgements}

We thank Sarthak Jain for pointing out an error in our original proof of what was Proposition 1, which led to the updated proposition and discussion in this version of the paper.

\bibliographystyle{acl_natbib}
\bibliography{citations}

\clearpage

\appendix

\begin{table}[t!]
\centering
\footnotesize
\begin{tabular}{p{0.45\linewidth}|p{0.45\linewidth}}
\toprule
\multicolumn{1}{c|}{Head: \textbf{\textit{Noun}}} &\multicolumn{1}{c}{Head: \textbf{\textit{Verb}}}\\\midrule
I think 2012 is going to be a great year for Fujairah as we have A LOT of projects to be done by 2012. & Went to the Willow Lounge this past weekend for dinner and drinks ... place is awesome.\\\\
V: \textit{going} & V: \textit{Went}\\
 NP: \textit{year} & NP: \textit{weekend}\\
PP: \textit{for Fujairah} & PP: \textit{for dinner and drinks}\\
\bottomrule
\end{tabular}
\caption{Examples of constructions extracted from the UD EWT training data using the heuristics described in Appendix \ref{sec:tuple-extraction}. Shown are sentences with differing PP attachment types for tuple (\textit{go, for}).}
\label{tab:tuple-examples}
\end{table}

\section{Ambiguous PP Attachment Extraction}
\label{sec:tuple-extraction}
Here, we describe how we heuristically extract (verb, noun, prepositional phrase) constructions with ambiguous attachment from the UD English Web Treebank training data (Section \ref{sec:empirical-data}). Examples of such constructions are shown in Table \ref{tab:tuple-examples}. We extract (V, N, PP) constructions from UD EWT inputs that meet the following criteria, which operate over the dependency relation annotations:
\begin{enumerate}
    \item V, NP, and PP are contained in same sentence
    \item \textit{Either} PP depends on NP \textit{or} PP depends on V
    \item NP depends on V and is not subject of V
    \item PP follows both V and NP in the sentence
\end{enumerate}

\section{Proof of Proposition 1}
\label{sec:prop1-proof}

This section will rely on the assumption of pairwise independence between input features, i.e., $x_i, x_j$ are independent for all $i.j$.

\begin{lemma}
Assume input features are pairwise independent. Then, $p_b(y \mid \neg x_i) = \frac{1}{2}$.
\begin{proof}
Let $f_i = p_u(x_i)$. We first derive the joint distribution $p_b(y, \neg x_i)$:
\begin{align*}
    p_b(y, \neg x_i) &= \frac{1}{2}(1 - f_i) + \frac{1}{2}f_ir_i p_b(y, \neg x_i) \\
    2p_b(y, \neg x_i) &= 1 - f_i + f_i r_i p_b(y, \neg x_i) \\
    p_b(y, \neg x_i) &= \frac{1 - f_i}{2 - f_i r_i} .
\end{align*}
We now derive the marginal probability $p_b(\neg x_i)$:
\begin{align*}
    p_b(\neg x_i) &= 1 - f_i + \frac{1}{2}f_i r_i p_b(\neg x_i) \\
    2p_b(\neg x_i) &= 2 - 2f_i + f_i r_i p_b(\neg x_i) \\
    p_b(\neg x_i) &= \frac{2(1 - f_i)}{2 - f_i r_i} .
\end{align*}
Now, we compute $p_b(y \mid \neg x_i)$:
\begin{align*}
    p_b(y \mid \neg x_i)
        = \frac{p_b(y, \neg x_i)}{p_b(\neg x_i)}
        = \frac{1}{2} .
\end{align*}
\end{proof}
\end{lemma}
We now turn to the main proof of \autoref{thm:edits}. Recall that we define the edit sensitivity $s_i$ of feature $i$ as $p(y' = \neg y \mid x'_i)$.
We also similarly define $e_i = p(x'_i = \neg x_i \mid x'_i)$.

\paragraph{Proposition 1.}
\textit{Assume $x_i$, $x_j$, $e_i$, $e_j$, $s_i$, and $s_j$ are independent for all $i$, $j$.
Then $p_e(y' \mid x'_i) = \frac{1}{2}$ if and only if $r_i = 0$ or $s_i = \frac{1}{2}$ or $e_i = 1$.}
\begin{proof}
We first consider the case where $y' = y$, and derive $p(y \mid x'_i)$. Let $e_i = p(\neg x_i \mid x'_i)$.
\begin{align*}
    p(y \mid x'_i)
        &= p(y \mid x_i) p(x_i \mid x'_i) \\
        &\quad+ p(y \mid \neg x_i) p(\neg x_i \mid x'_i) \\
        &= \frac{1}{2 - r_i} (1-e_i) + \frac{1}{2} e_i \\
        &= \frac{2(1 - e_i) + e_i(2-r_i)}{2(2 - r_i)} \\
        &= \frac{2 - e_i r_i}{2(2 - r_i)} .
\end{align*}
Now we let $y' = \neg y$ and derive $p_b(\neg y \mid x'_i)$:
\begin{align*}
    p(\neg y \mid x'_i)
        &= p(\neg y \mid x_i) p_b(x_i \mid x'_i) \\
        &\quad + p(\neg y \mid \neg x_i) p(\neg x_i \mid x'_i) \\
        &= \frac{1 - r_i}{2 - r_i} (1-e_i) + \frac{1}{2} e_i \\
        &= \frac{2 - 2r_i + e_ir_i}{2(2 - r_i)} .
\end{align*}
Finally, we write out $p(y' \mid x'_i)$ as
\begin{align*}
    p(y' \mid x'_i)
        &= p(y \mid x'_i) p(y'=y \mid x'_i) \\
        &\quad + p(\neg y \mid x'_i) p(y' \neq y \mid x'_i) \\
        &= \frac{2 - e_i r_i}{2(2 - r_i)} (1 - s_i) + \frac{2 - 2r_i + e_ir_i}{2(2 - r_i)} s_i \\
        &= \frac{2 - 2r_i s_i + r_i e_i(2s_i - 1)}{2(2-r_i)} \\
        &= \frac{1 - r_i(s_i - e_is_i + \frac{e_i}{2})}{2 - r_i} .
\end{align*}
From here, we set $p(y' \mid x'_i)$ to $\frac{1}{2}$ to prove the forward direction of our proposition.  The reverse direction can be easily verified by substituting the solutions found below back into the above equation.
\begin{align*}
    \frac{1 - r_i(s_i - e_is_i + \frac{e_i}{2})}{2 - r_i} &= \frac{1}{2} \\
    2 - 2r_i(s_i - e_is_i + \frac{e_i}{2}) &= 2 - r_i \\
    r_i(1 - 2(s_i - e_is_i + \frac{e_i}{2})) &= 0 \\
    r_i ( 1 - 2s_i + 2e_is_i - e_i) &= 0 \\
    r_i((1-e_i)-2s_i(1-e_i)) &= 0 \\
    r_i(1-e_i)(1-2s_i) &= 0
\end{align*}
Interestingly, this equation factorizes into three independent solutions, giving three ways to achieve an unbiased $p_e(y'\mid x')$: $r_i = 0$, $e_i = 1$, and $s_i = \frac{1}{2}$.  The implications of these solutions are discussed in the main text.
\end{proof}

\end{document}